\pdfoutput=1

\documentclass[11pt]{article}
\usepackage[final]{acl}
\usepackage{times}
\usepackage{latexsym}
\usepackage[T1]{fontenc}
\usepackage[utf8]{inputenc}
\usepackage{microtype}
\usepackage{inconsolata}
\usepackage{graphicx}

\usepackage{soul}

\let\origfootnote\footnote
\renewcommand{\footnote}[1]{\kern.1em\origfootnote{#1}}
\newcommand{\punctfootnote}[1]{\kern-.1em\origfootnote{#1}}

\definecolor{bottlegreen}{rgb}{0.0,0.42,0.31}

\usepackage[normalem]{ulem}
\def\ODdel#1{\bgroup\markoverwith{\textcolor{blue!60!green!100!black}{\rule[0.4ex]{2pt}{3pt}}}\ULon{#1}}

\def\tomaszdel#1{\bgroup\markoverwith{\textcolor{bottlegreen}{\rule[0.4ex]{2pt}{3pt}}}\ULon{#1}}

\title{Teaching LLMs at Charles University: Assignments and Activities}

\author{Jindřich Helcl \and Zdeněk Kasner \and Ondřej Dušek \and Tomasz Limisiewicz  \\ \and {\bf Dominik Macháček} \and {\bf Tomáš Musil} \and {\bf Jindřich Libovický} \\
        Institute of Formal and Applied Linguistics \\ 
        Faculty of Mathematics and Physics, Charles University \\
        V Holešovičkách 2, 180 00, Prague, Czech Republic \\
        \texttt{\$\{surname\}@ufal.mff.cuni.cz}}

\begin{document}
\maketitle
\begin{abstract}
This paper presents teaching materials, particularly assignments and ideas for classroom activities, from a new course on large language models
(LLMs) taught at Charles University.
The assignments include experiments with LLM inference for weather report generation and machine translation.
The classroom activities include class quizzes, focused research on downstream tasks and datasets, and an interactive "best paper" session aimed at reading and comprehension of research papers.
\end{abstract}

\section{Introduction}

Reflecting contemporary trends in education is a challenging task.
The teachers often need to decide which promising topics to cover in their course and which topics are better to leave for discussion in reading groups. The unstable nature
of research progress also means that courses that are not updated regularly lose their
relevance in time. However, when the trend becomes as prominent as large language models (LLMs) have become, the need for a systematic overview in the form of a specialized course gets increasingly urgent.

This paper presents one of these efforts -- a new course taught at Charles University composed of a series of LLM-related lectures and interactive sessions. During its first year in 2024, 51 students enrolled in the optional course, mainly attending
local BSc or MSc study programmes.



The course is composed of 13 sessions with various levels of interactivity, including lectures, directed discussions on the current topic, quizzes, as well as practical work with LLMs (Section~\ref{sec:practical}) and other classroom activities (Section~\ref{sec:activities}). After 
a broader discussion in the first session, the course focused on the following topics: The Transformer model \citep{vaswani2017attention}, LLM training and inference, data collection and evaluation, LLM applications, efficiency, multilinguality, speech processing, translation, meaning/understanding, and ethics of LLM training and use.


All course materials, including slides, recordings, and assignments, are available on the course website.\punctfootnote{\url{https://ufal.mff.cuni.cz/courses/npfl140}}

\section{Assignments}
\label{sec:practical}

We organized two assignment-based sessions focused on (1) generating weather reports using LLMs and (2) using LLMs for machine translation (MT). For each task, we ran instances of different models on our GPU cluster with an API provided by the \texttt{text-generation-webui}\footnote{\url{https://github.com/oobabooga/text-generation-webui}} package. The API allowed the students to access and configure the models without the need to access specialized hardware or rely on commercial platforms.

In both tasks, the students worked in small teams (up to 5 people), and were provided with a starter code\footnote{\url{https://github.com/kasnerz/npfl140}} that would call the model API with a specified set of parameters. Besides choosing the appropriate prompts, the teams experimented with various decoding parameters, including the sampling temperature, the $k$ and $p$ parameters for top-$k$ and top-$p$ sampling, and the beam size.

\paragraph{Text Generation.}
In this task, the students were asked to generate weather reports in natural language using an LLM. The students were provided with a selection of JSON files retrieved from the \url{openweathermap.org} API for various cities.
The assignment was divided into 4 subtasks: (1) generating a report about the current weather, (2) generating a 5-day forecast, (3) generating a report in a language other than English, and (4) changing the forecast style (e.g., for specific target groups).

The four models the students experimented with were Mistral 7B,\punctfootnote{\url{https://hf.co/mistralai/Mistral-7B-v0.1}} Mistral 7B Instruct \citep{jiang2023mistral},\punctfootnote{\url{https://hf.co/mistralai/Mistral-7B-Instruct-v0.1}} Phi-2 \citep{phi2},\punctfootnote{\url{https://hf.co/microsoft/phi-2}} and Aya-101 \citep{ustun2024aya}.\punctfootnote{\url{https://hf.co/CohereForAI/aya-101}}
The students reported on the difficulties of generating factually accurate outputs from the models, confirming recent findings \cite{kasner2024beyond}. They also proposed improved data preprocessing, prompt formatting, and decoding parameters.

\paragraph{Machine Translation.}
In the MT assignment, the teams were given paragraphs of text
in 21 (unknown) languages and instructed to translate them using an LLM into English and then into a language of their choice. Again, the students experimented with prompt engineering and decoding parameters.

For the first part, we created a simple web app\footnote{\url{https://github.com/jlibovicky/llm-mt-assignment}}
for submitting the English translation, which computed the Character F-score \citep{popovic-2015-chrf} and showed a leaderboard of the 10 best-scoring teams per language during the session. After the assignment, the leaderboard can be configured to show the source language and the reference translations. The leaderboard then shows all submissions made to the app except those marked as debug submissions. In the second part of the assignment, the students were asked to experiment with translation into a language of their choice. They should submit a report, which is due a week after the hands-on session.

We used a slightly different set of models compared to the previous assignment. Mistral 7B Instruct and Aya-101 remained, and we added translation-specific models Tower Instruct \citep{tower2024} and ALMA-R \citep{almar2024}.

When translating into English from medium-resourced languages, the students could generally match the quality of commercial MT systems. However, translating into their languages (e.g., Slovak, Ukrainian, Georgian, Serbian) appeared challenging.

\section{Classroom Activities}
\label{sec:activities}


\paragraph{Discussions.} 


Discussion among the students was a recurring activity in many of the sessions. To encourage as many students to participate, we either let them discuss in small groups and then present their position, or we used interactive slides to collect and show their input in real-time,\punctfootnote{Slido: \url{www.slido.com}} which also encouraged less self-confident students to share their opinion.

Discussion is an effective method for teaching non-technical topics. In the final session on this course, we focused on two primary areas. The first area involves the question of whether LLMs can truly understand language. We recommend engaging students in discussions about various thought experiments \citep[e.g.][]{searle, bender-2020} and exploring both sides of the debate: those who argue that it is impossible \citep[e.g.][]{parrots} and those who believe it is possible to some extent \citep[e.g.][]{andreas, sogaard}.
The second area covers ethical considerations. Here, students discussed environmental and labor issues related to training LLMs \citep[e.g.][]{parrots} and the broader challenges associated with the development and deployment of language technologies \citep[e.g.][]{jorgensen2023rawlsian}.


\paragraph{Class Quizzes.} 
Every session began with a short multiple-choice quiz based on the topics from the previous class. These quizzes were implemented using a simple web app\footnote{\url{https://github.com/jlibovicky/class-quiz}} that shows a QR code to join the quiz, and after a certain amount of time, it shows the results. Each question can be answered multiple times until the correct choice is selected, providing immediate feedback to the students. When the time is up, the app shows the correct answers and the number of unsuccessful attempts for each incorrect choice.

At the final session, students complete a similar immediate-feedback test in the form of scratch cards \citep{epstein2001immediate}.

\paragraph{Downstream tasks and datasets.}

During the class on LLM fine-tuning, we asked the students to split into groups and assigned downstream tasks (summarization, code generation, hate speech detection, and machine translation). The students were supposed to find suitable datasets and evaluation metrics. The groups presented their findings to the class and then discussed the potential drawbacks of the benchmarks and evaluation metrics.

\paragraph{Reading research papers.}

One of the goals we set for the course was to teach the students to responsibly assess the quality and trustworthiness of recent research papers. We organized an activity where the students role-played a best-paper committee, partially inspired by the Role-Playing Paper-Reading Seminars \citep{jacobson2021role}.

We selected five papers to encourage critical assessment of the values of model descriptions \citep{jiang2023mistral, schick2024toolformer} and analytical works \citep{basmova2023blind, balloccu-etal-2024-leak, dongkeun2024langbridge}. Each student was randomly assigned one of the papers to read thoroughly.



During the class, we first divided the students into groups containing at least one student per article, where the students explained the papers to each other. Then, the students re-grouped by their assigned article, where they discussed the paper again and nominated an advocate for and against it. Then, the advocates presented their final one-minute speeches. Finally, students secretly voted for the best paper using an online form.

\section{Conclusion}
We presented teaching materials and class activities for a new LLM course taught at Charles University. In the first year, 51 students enrolled course of which around 30 were actively participating. We expect a larger attendance in the following years after the course is upgraded from optional to elective.\punctfootnote{Enrolling in a subset of elective courses is mandatory in the study programme.}  All classroom activities can be applied in larger cohorts as well. Scaling up the LLM-based assignments for larger number of students might pose an issue for institutions with limited access to computing resources. However, we used four model setups (for each we needed one GPU) that were available for all the students, and so we used up only a relatively small portion of the resources available in our GPU cluster.
We therefore expect no severe issues with scaling up to a few hundreds of active students. The availability of enough teaching assistants to ensure proper feedback to the students will potentially become a more significant issue.




\section*{Acknowledgments}

We thank our colleagues Josef Jon, Peter Polák, and Rudolf Rosa, who helped to teach the course. Additional thanks to David Mareček, Michal Novák, Martin Popel, and Ondřej Plátek, who were involved in the course preparation.
The preparation of the course was partially supported by the Charles University project PRIMUS/23/SCI/023.


\bibliography{custom}

\begin{thebibliography}{20}
\providecommand{\natexlab}[1]{#1}

\bibitem[{Alves et~al.(2024)Alves, Pombal, Guerreiro, Martins, Alves, Farajian, Peters, Rei, Fernandes, Agrawal, Colombo, de~Souza, and Martins}]{tower2024}
Duarte~M. Alves, Jos{\'{e}} Pombal, Nuno~Miguel Guerreiro, Pedro~Henrique Martins, Jo{\~{a}}o Alves, M.~Amin Farajian, Ben Peters, Ricardo Rei, Patrick Fernandes, Sweta Agrawal, Pierre Colombo, Jos{\'{e}} G.~C. de~Souza, and Andr{\'{e}} F.~T. Martins. 2024.
\newblock \href {https://doi.org/10.48550/ARXIV.2402.17733} {Tower: An open multilingual large language model for translation-related tasks}.
\newblock \emph{CoRR}, abs/2402.17733.

\bibitem[{Andreas(2022)}]{andreas}
Jacob Andreas. 2022.
\newblock \href {https://doi.org/10.18653/v1/2022.findings-emnlp.423} {Language models as agent models}.
\newblock In \emph{Findings of the Association for Computational Linguistics: EMNLP 2022}, pages 5769--5779, Abu Dhabi, United Arab Emirates. Association for Computational Linguistics.

\bibitem[{Balloccu et~al.(2024)Balloccu, Schmidtov{\'a}, Lango, and Dusek}]{balloccu-etal-2024-leak}
Simone Balloccu, Patr{\'\i}cia Schmidtov{\'a}, Mateusz Lango, and Ondrej Dusek. 2024.
\newblock \href {https://aclanthology.org/2024.eacl-long.5} {Leak, cheat, repeat: Data contamination and evaluation malpractices in closed-source {LLM}s}.
\newblock In \emph{Proceedings of the 18th Conference of the European Chapter of the Association for Computational Linguistics (Volume 1: Long Papers)}, pages 67--93, St. Julian{'}s, Malta. Association for Computational Linguistics.

\bibitem[{Basmova et~al.(2023)Basmova, Goldberg, and Tsarfaty}]{basmova2023blind}
Victoria Basmova, Yoav Goldberg, and Reut Tsarfaty. 2023.
\newblock \href {https://doi.org/10.48550/ARXIV.2305.14785} {Chatgpt and simple linguistic inferences: Blind spots and blinds}.
\newblock \emph{CoRR}, abs/2305.14785.

\bibitem[{Bender et~al.(2021)Bender, Gebru, McMillan-Major, and Shmitchell}]{parrots}
Emily~M. Bender, Timnit Gebru, Angelina McMillan-Major, and Shmargaret Shmitchell. 2021.
\newblock \href {https://doi.org/10.1145/3442188.3445922} {On the dangers of stochastic parrots: Can language models be too big?}
\newblock In \emph{Proceedings of the 2021 ACM Conference on Fairness, Accountability, and Transparency}, FAccT '21, page 610–623, New York, NY, USA. Association for Computing Machinery.

\bibitem[{Bender and Koller(2020)}]{bender-2020}
Emily~M. Bender and Alexander Koller. 2020.
\newblock \href {https://doi.org/10.18653/v1/2020.acl-main.463} {Climbing towards {NLU}: {On} meaning, form, and understanding in the age of data}.
\newblock In \emph{Proceedings of the 58th Annual Meeting of the Association for Computational Linguistics}, pages 5185--5198, Online. Association for Computational Linguistics.

\bibitem[{Epstein et~al.(2001)Epstein, Epstein, and Brosvic}]{epstein2001immediate}
Michael~L Epstein, Beth~B Epstein, and Gary~M Brosvic. 2001.
\newblock Immediate feedback during academic testing.
\newblock \emph{Psychological reports}, 88(3):889--894.

\bibitem[{Jacobson and Raffel(2021)}]{jacobson2021role}
Alec Jacobson and Colin Raffel. 2021.
\newblock \href {https://www.cs.toronto.edu/~jacobson/images/role-playing-paper-reading-seminars.pdf} {Role-playing paper-reading seminars}.
\newblock {\it Colin Raffel's Blog}.

\bibitem[{Javaheripi et~al.(2023)Javaheripi, Bubeck, Abdin, Aneja, C\'{e}sar, Mendes, Chen, Giorno, Eldan, Gopi, Gunasekar, Kauffmann, Lee, Li, Nguyen, de~Rosa, Saarikivi, Salim, Shah, Santacroce, Behl, Kalai, Wang, Ward, Witte, Zhang, and Zhang}]{phi2}
Mojan Javaheripi, S\'{e}bastien Bubeck, Marah Abdin, Jyoti Aneja, Caio C\'{e}sar, Teodoro Mendes, Weizhu Chen, Allie~Del Giorno, Ronen Eldan, Sivakanth Gopi, Suriya Gunasekar, Piero Kauffmann, Yin~Tat Lee, Yuanzhi Li, Anh Nguyen, Gustavo de~Rosa, Olli Saarikivi, Adil Salim, Shital Shah, Michael Santacroce, Harkirat~Singh Behl, Adam~Taumann Kalai, Xin Wang, Rachel Ward, Philipp Witte, Cyril Zhang, and Yi~Zhang. 2023.
\newblock \href {https://www.microsoft.com/en-us/research/blog/phi-2-the-surprising-power-of-small-language-models} {Phi-2: The surprising power of small language models}.
\newblock {\it Microsoft Research Blog}.

\bibitem[{Jiang et~al.(2023)Jiang, Sablayrolles, Mensch, Bamford, Chaplot, de~las Casas, Bressand, Lengyel, Lample, Saulnier, Lavaud, Lachaux, Stock, Scao, Lavril, Wang, Lacroix, and Sayed}]{jiang2023mistral}
Albert~Q. Jiang, Alexandre Sablayrolles, Arthur Mensch, Chris Bamford, Devendra~Singh Chaplot, Diego de~las Casas, Florian Bressand, Gianna Lengyel, Guillaume Lample, Lucile Saulnier, Lélio~Renard Lavaud, Marie-Anne Lachaux, Pierre Stock, Teven~Le Scao, Thibaut Lavril, Thomas Wang, Timothée Lacroix, and William~El Sayed. 2023.
\newblock \href {https://arxiv.org/abs/2310.06825} {Mistral 7b}.
\newblock \emph{Preprint}, arXiv:2310.06825.

\bibitem[{J{\o}rgensen and S{\o}gaard(2023)}]{jorgensen2023rawlsian}
Anna~Katrine J{\o}rgensen and Anders S{\o}gaard. 2023.
\newblock Rawlsian ai fairness loopholes.
\newblock \emph{AI and Ethics}, 3(4):1185--1192.

\bibitem[{Kasner and Du{\v{s}}ek(2024)}]{kasner2024beyond}
Zden{\v{e}}k Kasner and Ond{\v{r}}ej Du{\v{s}}ek. 2024.
\newblock Beyond reference-based metrics: Analyzing behaviors of open llms on data-to-text generation.
\newblock \emph{arXiv preprint arXiv:2401.10186}.

\bibitem[{Popovi{\'c}(2015)}]{popovic-2015-chrf}
Maja Popovi{\'c}. 2015.
\newblock \href {https://doi.org/10.18653/v1/W15-3049} {chr{F}: character n-gram {F}-score for automatic {MT} evaluation}.
\newblock In \emph{Proceedings of the Tenth Workshop on Statistical Machine Translation}, pages 392--395, Lisbon, Portugal. Association for Computational Linguistics.

\bibitem[{Schick et~al.(2023)Schick, Dwivedi{-}Yu, Dess{\`{\i}}, Raileanu, Lomeli, Hambro, Zettlemoyer, Cancedda, and Scialom}]{schick2024toolformer}
Timo Schick, Jane Dwivedi{-}Yu, Roberto Dess{\`{\i}}, Roberta Raileanu, Maria Lomeli, Eric Hambro, Luke Zettlemoyer, Nicola Cancedda, and Thomas Scialom. 2023.
\newblock \href {http://papers.nips.cc/paper\_files/paper/2023/hash/d842425e4bf79ba039352da0f658a906-Abstract-Conference.html} {Toolformer: Language models can teach themselves to use tools}.
\newblock In \emph{Advances in Neural Information Processing Systems 36: Annual Conference on Neural Information Processing Systems 2023, NeurIPS 2023, New Orleans, LA, USA, December 10 - 16, 2023}.

\bibitem[{Searle(1985)}]{searle}
John~R. Searle. 1985.
\newblock \emph{Minds, brains, and programs}, page 282–307.
\newblock MIT Press, Cambridge, MA, USA.

\bibitem[{S{\o}gaard(2022)}]{sogaard}
Anders S{\o}gaard. 2022.
\newblock Understanding models understanding language.
\newblock \emph{Synthese}, 200(6):443.

\bibitem[{Vaswani et~al.(2017)Vaswani, Shazeer, Parmar, Uszkoreit, Jones, Gomez, Kaiser, and Polosukhin}]{vaswani2017attention}
Ashish Vaswani, Noam Shazeer, Niki Parmar, Jakob Uszkoreit, Llion Jones, Aidan~N Gomez, {\L}ukasz Kaiser, and Illia Polosukhin. 2017.
\newblock \href {https://proceedings.neurips.cc/paper_files/paper/2017/file/3f5ee243547dee91fbd053c1c4a845aa-Paper.pdf} {Attention is all you need}.
\newblock In \emph{Advances in Neural Information Processing Systems}, volume~30. Curran Associates, Inc.

\bibitem[{Xu et~al.(2024)Xu, Sharaf, Chen, Tan, Shen, Durme, Murray, and Kim}]{almar2024}
Haoran Xu, Amr Sharaf, Yunmo Chen, Weiting Tan, Lingfeng Shen, Benjamin~Van Durme, Kenton Murray, and Young~Jin Kim. 2024.
\newblock \href {https://doi.org/10.48550/ARXIV.2401.08417} {Contrastive preference optimization: Pushing the boundaries of {LLM} performance in machine translation}.
\newblock \emph{CoRR}, abs/2401.08417.

\bibitem[{Yoon et~al.(2024)Yoon, Jang, Kim, Kim, Shafayat, and Seo}]{dongkeun2024langbridge}
Dongkeun Yoon, Joel Jang, Sungdong Kim, Seungone Kim, Sheikh Shafayat, and Minjoon Seo. 2024.
\newblock \href {https://doi.org/10.48550/ARXIV.2401.10695} {Langbridge: Multilingual reasoning without multilingual supervision}.
\newblock \emph{CoRR}, abs/2401.10695.

\bibitem[{Üstün et~al.(2024)Üstün, Aryabumi, Yong, Ko, D'souza, Onilude, Bhandari, Singh, Ooi, Kayid, Vargus, Blunsom, Longpre, Muennighoff, Fadaee, Kreutzer, and Hooker}]{ustun2024aya}
Ahmet Üstün, Viraat Aryabumi, Zheng-Xin Yong, Wei-Yin Ko, Daniel D'souza, Gbemileke Onilude, Neel Bhandari, Shivalika Singh, Hui-Lee Ooi, Amr Kayid, Freddie Vargus, Phil Blunsom, Shayne Longpre, Niklas Muennighoff, Marzieh Fadaee, Julia Kreutzer, and Sara Hooker. 2024.
\newblock Aya model: An instruction finetuned open-access multilingual language model.
\newblock \emph{arXiv preprint arXiv:2402.07827}.

\end{thebibliography}

\end{document}